\documentclass{kais}
\usepackage{wrapfig,epsf}
\usepackage{natbib,har2nat}
\usepackage{graphicx} 
\usepackage{paralist}
\usepackage[english]{babel}
\usepackage{booktabs}
\usepackage{url}
\usepackage{commath}
\usepackage{amsmath}
\usepackage{amsfonts}
\usepackage{amssymb}
\usepackage{mathtools}
\newcommand\myeq{\stackrel{\mathclap{\normalfont\mbox{def}}}{=}}

\bibpunct[, ]{(}{)}{;}{a}{}{,}

\newtheorem{defi}{Definition}[section]

\received{22 Feb 2021}
\revised{11 Aug 2021}
\accepted{05 Dec 2021}

\pubyear{2021}
\pagerange{\pageref{firstpage}--\pageref{lastpage}}
\volume{xxx}

\begin{document}
\label{firstpage}

\title{NeuCrowd: Neural Sampling Network for Representation Learning with Crowdsourced Labels}

\author[Y. Hao et al]{Yang Hao, Wenbiao Ding and Zitao Liu\\ TAL Education Group, Beijing, China}
 
\maketitle

\begin{abstract}
Representation learning approaches require a massive amount of discriminative training data, which is unavailable in many scenarios, such as healthcare, smart city, education, etc. In practice, people refer to crowdsourcing to get annotated labels. However, due to issues like data privacy, budget limitation, shortage of domain-specific annotators, the number of crowdsourced labels is still very limited. Moreover, because of annotators' diverse expertise, crowdsourced labels are often inconsistent. Thus, directly applying existing supervised representation learning (SRL) algorithms may easily get the overfitting problem and yield suboptimal solutions. In this paper, we propose \emph{NeuCrowd}, a unified framework for SRL from crowdsourced labels. The proposed framework (1) creates a sufficient number of high-quality \emph{n}-tuplet training samples by utilizing safety-aware sampling and robust anchor generation; and (2) automatically learns a neural sampling network that adaptively learns to select effective samples for SRL networks. The proposed framework is evaluated on both one synthetic and three real-world data sets. The results show that our approach outperforms a wide range of state-of-the-art baselines in terms of prediction accuracy and AUC. To encourage reproducible results, we make our code publicly available at \url{https://github.com/tal-ai/NeuCrowd_KAIS2021}.

\end{abstract}

\begin{keywords}
Crowdsourcing; Representation learning; Sampling network
\end{keywords}

\section{Introduction}
\label{sec:intro}
Representation learning, especially deep learning, has been proven to be crucial in many different domains such as information retrieval \cite{grbovic2018real}, recommender systems \cite{xue2017deep}, computer vision \cite{duan2019deep,sohn2016improved}, etc. Such approaches are usually discriminatively trained on massive labeled data sets, which are mostly generated from explicit or implicit online user engagement, like ratings, comments, clicks, and hides \cite{bengio2013representation}. 

However, in many real-world scenarios such as healthcare, smart city, education, finance, etc., labeled data sets are typically insufficient or unavailable. To alleviate this problem, human efforts can be involved to acquire labeled data manually and crowdsourcing provides a flexible solution \cite{whitehill2009whose,raykar2010learning,rodrigues2014gaussian,soto2017crowdsourcing}. Theoretically, we could annotate data sets as large as we want via crowdsourcing platforms such as Amazon Mechanical Turk \footnote{https://www.mturk.com/}, CrowdTruth\footnote{http://crowdtruth.org/}, etc. Unfortunately, the number of crowdsourced labels is still very limited due to a variety of reasons as follows: 

\begin{itemize}
\item \textbf{data privacy}: data sets in many offline scenarios are difficult to collect due to privacy concerns. For example, in medical diagnostic imaging, patient data sets are prohibited to the public by applicable laws \cite{price2019privacy,orgill2004urgency,rodriguez2016learning}.
\item \textbf{specialist shortage}: crowdsourced tasks may require domain specialties. For instance, in educational data mining, student assessments require pedagogical specialties from annotators, which doesn't scale by nature \cite{kittur2008crowdsourcing,schenk2011towards}.
\item \textbf{high cost}: labeling tasks may require excessive budgets or tedious efforts. For example, it may take a crowd worker less than 1 second to annotate an image while a worker has to watch a 60-min classroom recording before determining the class quality, i.e., whether the class is good or bad \cite{chen2019multimodal}.
\end{itemize}
 
Recent years have witnessed great efforts on learning with small labeled data \cite{fei2006one,wang2020representation,ravi2016optimization,vinyals2016matching}. Meanwhile inferring true labels from inconsistent crowdsourced labels has been studied for decades \cite{whitehill2009whose,raykar2010learning,rodrigues2014gaussian,li2021crowdrl,hao2021temporal}. However, research on supervised representation learning (SRL) with small and inconsistent crowdsourced labels is rather limited. Therefore, the objective of this work is to study and develop approaches that can be used for learning representation from crowdsourced labels. More specifically, we target on answering two questions: (1) since annotated samples in healthcare, education and many other domains are usually in an incredibly smaller order of magnitude (a few hundred or less), compared to web-scale data sets, how do we take advantage of deep representation learning under the limited sample setting? and (2) due to the fact that crowdsourced labels may be highly inconsistent, how do we handle such uncertainty and make the learning procedure more efficient?

In this work, we address the above issues by presenting a unified framework \emph{NeuCrowd} that is applicable to learn effective representations from very limited crowdsourced data. We propose a scheme of generating hundreds of thousands of safety-aware and robust training instances from a limited amount of inconsistent crowdsourced labeled data.

Our data augmentation approach generalizes the deep triplet embedding learning in computer vision into crowdsourcing settings with multiple negative examples, a.k.a., \emph{n}-tuplet, where each \emph{n}-tuplet consists of an anchor, a positive example, and \emph{n-2} negative examples \cite{sohn2016improved,xu2019learning}. Furthermore, in order to expedite the learning process and improve the quality of the learned representations, we specifically design a neural sampling network to adaptively select ``hard'' \emph{n}-tuplet training samples. Different from most existing hard example mining heuristics \cite{shrivastava2016training}, our framework is able to train both the representation learning network and the sampling network simultaneously. Hence, the sampling network is able to dynamically exploit relations among  \emph{n}-tuplet samples without any hard-coding heuristic.

Overall this paper makes the following contributions:

\begin{itemize}
\item We propose a safety-aware and robust data augmentation technique that considers the inconsistency and uncertainty between examples and creates a sufficient number of robust \emph{n}-tuplet training samples.

\item We design a sampling network to automatically and adaptively select optimized (a.k.a., hard) \emph{n}-tuplet samples for the representation learning framework. The sampling network doesn't rely on any pre-fixed heuristic and both the embedding network and the sampling network are optimized simultaneously.

\item We conduct a detailed and comprehensive experimental comparison of the proposed framework on multiple data sets from different domains. To encourage reproducible results, we make our code and data publicly available on a github repository.
\end{itemize}

\section{Related Work}
\label{sec:related}

\subsection{Learning with Limited Data}

Both few/zero-shot learning and semi/weakly supervised learning approaches have been developed to enable learning with limited labeled data in different ways. Motivated by the fact that humans can learn new concepts with very little supervision,  few/zero-shot learning models aim to learn new concepts from a very small number of labeled examples \cite{fei2006one,snell2017prototypical,sung2018learning}. While semi/weakly supervised learning approaches make use of a large amount of unlabeled data to learn better predictors \cite{takamatsu2012reducing,ratner2016data}.

Although few-shot learning methods yield promising results on unseen categories, they demand large data sets from other categories. This may be infeasible in many real-world domains other than computer vision. Similarly, semi-supervised or weekly supervised approaches, may not work when the total available data is limited.


\subsection{Learning with Crowdsourced Labels}

Truth inference is well studied in crowdsourcing research \cite{whitehill2009whose,raykar2010learning,rodrigues2014gaussian}, which aims at directly inferring the ground truth from workers' annotations. \citeasnoun{whitehill2009whose} proposed a probabilistic framework that iteratively adjusts the inferred ground truth estimates based on the performance of the annotators. \citeasnoun{raykar2010learning} proposed an EM algorithm to jointly learn the levels of annotators and the regression models. \citeasnoun{rodrigues2014gaussian} generalized Gaussian process classification to consider multiple annotators with diverse expertise .


The majority of aforementioned algorithms have been designed to address the label inconsistency problem and they cannot work as expected when labels are limited. In this work, we aim to develop algorithms which can jointly solve the SRL challenges from limited and inconsistent labels. 

\subsection{Metric Learning with Hard Example Mining}

Deep metric learning approaches construct pairs \cite{koch2015siamese,sohn2016improved} or triplets \cite{schroff2015facenet,he2018triplet} with different objective functions. Consequently, various hard example mining techniques are developed to select ``hard'' training samples to expedite the optimization convergence \cite{sung1996learning}. Many approaches along this direction have achieved promising results in many tasks such as object detection \cite{shrivastava2016training}, face recognition \cite{sohn2016improved,schroff2015facenet}, etc.

Although deep metric learning approaches with hard example mining can learn effective representations, they require a large amount of data. Moreover, they heavily rely on the comparisons within pairs or triplets, which are very sensitive to ambiguous examples and may be easily misled by inconsistent crowdsourced labels.

\section{Methodology}
\label{sec:method} 
\subsection{Notation}

Following conventions, we use bold upper case for collections or sets, bold lower case letters for vectors and calligraphic typeface for functions. We use $(\cdot)^+$ and $(\cdot)^-$ to indicate positive and negative examples. More specifically, let $\mathbf{D}$ be the original crowdsourced data set, i.e., $\mathbf{D} = \{ e_i \}_{i=1}^N = \{(\mathbf{x}_i, \mathbf{y}_i)\}_{i=1}^N$, where $e_i$ is the \emph{i}th example and $\mathbf{x}_i$ denotes the raw features of $e_i$ and $\mathbf{y}_i \in \mathbb{R}^{d \times 1}$ denotes the crowdsourced labels of $e_i$. \emph{N} is the size of $\mathbf{D}$ and $d$ is the number of crowd workers. Let $y_{ij} \in \{0, 1\}$ be the binary label from the \emph{j}th worker for $e_i$, i.e., $\mathbf{y}_i = [y_{i1}, \cdots, y_{id}]$. 

\subsection{\emph{N}-tuplet}

Similar to \citeasnoun{sohn2016improved} and \citeasnoun{xu2019learning}, we define \emph{n}-tuplet as follows:

\begin{defi}{(\textsc{\emph{n}-tuplet})}
\label{def:tuplet}
An \emph{n}-tuplet $\mathbf{T}$ is an n-element collection that consists of an anchor, a positive example (to the anchor) and n-2 negative examples, i.e., 

$$\mathbf{T} \quad \myeq \quad (e^*, e^+, e^-_1, \cdots, e^-_{n-2} )$$

\noindent where $e^*$ is an anchor and $e^+$ is a positive example to $e^*$ and $\{e^-_i\}_{i=1}^{n-2}$ are negative examples.  An anchor is randomly chosen from all examples each time we construct an n-tuplet. Then a second example from the same class is chosen as a positive example, and examples chosen from other classes are negative. 

\end{defi}

The \emph{n}-tuplet is a generalization of triplet (when $n=3$) where it contains more than one negative examples. The \emph{n}-tuplet objective function allows joint comparisons among $n-2$ negative examples. In terms of model learning, different from triplets that only support learning from negative examples one at a time, the \emph{n}-tuplets try to maximize the distances between positive examples and all the other $n-2$ negative examples all at once \cite{sohn2016improved}. 

As we discussed, the limited number of labeled data in certain domains may easily lead to the overfitting problems for many SRL approaches. Fortunately, this issue can be largely alleviated by exploiting the trick of \emph{n}-tuplets \cite{sohn2016improved,xu2019learning}. Theoretically, we could create a size of $\mathcal{O}(P^2Q^{n-2})$ \emph{n}-tuplets where $P$ and $Q$ are the numbers of positive and negative examples. By sampling and reassembling from the original data set $\mathbf{D}$, we are able to significantly augment the training data size compared to the one that directly training models from individual examples, i.e., $\mathcal{O}(P+Q)$. Therefore, in this work, we develop our SRL framework that is optimized on batches of \emph{n}-tuplets instead of individual examples. 

\subsection{The NeuCrowd Model}

Although the \emph{n}-tuplet based paradigm is able to address the problem of learning from small data, building end-to-end SRL solutions from crowdsourced labels is still challenging and gives rise to the following important questions:

\begin{itemize}
\item[\emph{Q1}]: How do we effectively construct \emph{n}-tuplets from highly inconsistent crowdsourced data?
\item[\emph{Q2}]: How do we improve the efficiency of the embedding training when using a set of \emph{n}-tuplets?
\end{itemize}

\begin{figure}
\centering
\includegraphics[width=0.75\textwidth] {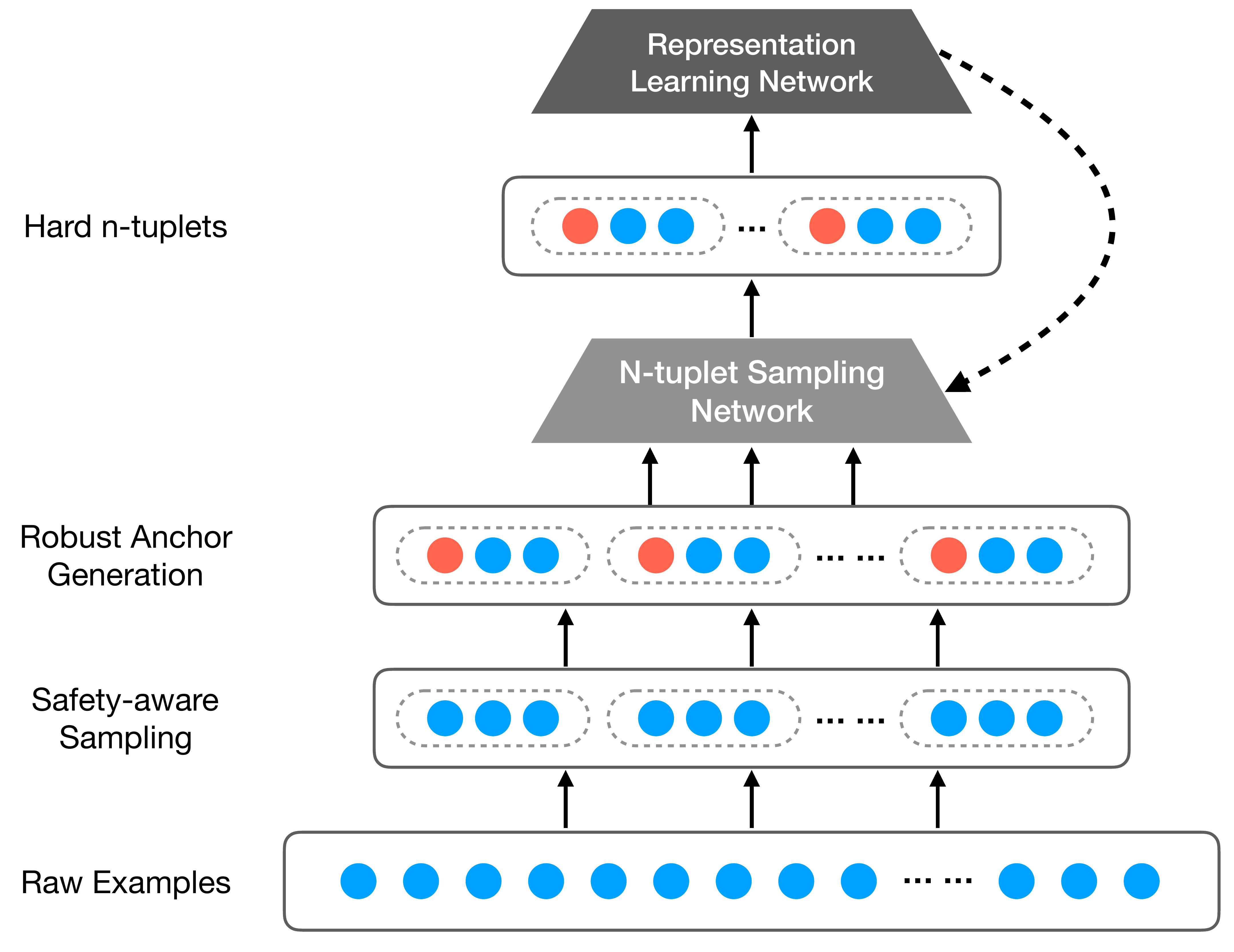}
\caption{The overview of the \emph{NeuCrowd} framework. The dash line indicates that the SRL network passes its training loss to the \emph{n}-tuplet sampling network. The blue dots represent original crowdsourced examples and red dots represent the robust anchors.}
\label{fig:model}
\end{figure}

In this work, we address \emph{Q1} by proposing (1) a safety-aware sampling strategy to ``clean up'' the \emph{n}-tuplet construction space by dynamically identifying inconsistent crowdsourced examples along with the embedding learning; and (2) a robust anchor generation method to artificially create anchors that reduce ambiguity and chances of outliers within \emph{n}-tuplets. To answer \emph{Q2}, we develop a deep embedding network that is able to learn from \emph{n}-tuplets and explicitly design a sampling network, which adaptively selects the ``hardest'' \emph{n}-tuplets and co-learns its parameters with the SRL network without any heuristic. The iterative joint learning paradigm is described in the ``Joint Learning Paradigm" Section. The entire \emph{NeuCrowd} framework is illustrated in Figure \ref{fig:model}.

\subsubsection{Safety-aware Sampling}
\label{sec:safe}

When obtaining reliable labels from the crowds, an example is usually annotated by multiple workers \cite{raykar2010learning,yan2014learning,yan2010modeling}. Consequentially, we may get different votes for the same example. For example, assuming that $e^+_i$ and $e^+_j$ are two positive examples whose corresponding 5-person crowdsourced binary labels are (1, 1, 1, 1, 1) and (1, 1, 1, 0, 0), our assurance of the labels of $e^+_i$ and $e^+_j$ should be different. Here we refer to label assurance as the measurement of the degree of disagreement of annotated labels within the same example, which is defined as follows:

\begin{defi}{(\textsc{Label Assurance})}
\label{def:assurance}
Given a crowdsourced example $e_i$, its label assurance, i.e., $\mathcal{A}(e_i)$ is computed as follows:

$$\mathcal{A}(e_i) = \abs{ 1 - \frac{2}{d} \|\mathbf{y_i}\|_1 }$$

\noindent where $\abs{\cdot}$ denotes the absolute value and $\|\cdot\|_1$ represents the vector $\ell_1$ norm.
\end{defi}

Our label assurance measures the disagreement degree among workers and reaches its minimum value\footnote{The minimum value goes to $1/d$ when $d$ is odd.} of 0 when a tie or a draw happens and goes to its maximum value of 1 when all labels are consistent. An equivalent approach is to compute the maximum likelihood estimation (MLE) of label  $y^{MLE} = \|\mathbf{y_i}\|_1/d$, similar to \citeasnoun {xu2019learning}, then $\mathcal{A}(e_i)$ is measured as the distance between MLE label and the decision threshold (0.5, by default) then re-scaled to 0-1:

$$\mathcal{A}(e_i) = 2\abs{y^{MLE}  - 0.5} = \abs{ 1 - \frac{2}{d} \|\mathbf{y_i}\|_1 }$$ 

\noindent where Bayesian estimation results can be substituted for $y^{MLE}$, if prior knowledge is accessible \cite{sheng2008get}.

Since the SRL network optimizes its parameters purely from \emph{n}-tuplets and it tries to push the $n-2$ negative examples all at once within each \emph{n}-tuplet, incorporating unsure labels will easily confuse the objective function and lead to inferior representations. Therefore, it is necessary to exclude those ambiguous examples when constructing the training set. Due to the fact that such ambiguous instances may make up 50\% of all labels, simply discarding all the ambiguous examples directly doesn't work \cite{zhong2015active,takeoka2020learning}. 

Therefore, we propose a safety-aware sampling technique to dynamically sample safe examples in the embedding space when constructing the \emph{n}-tuplets. Here at each training iteration \emph{t}, we define the safe example as follows:

\begin{defi}{(\textsc{Safe Example})}
\label{def:safety}
Let $\mathbf{N}_i(t)^+$ and $\mathbf{N}_i(t)^-$ denote the index sets of k nearest neighbors for $e_i$ at iteration t in the embedding space, where indices in $\mathbf{N}_i(t)^+$ belong to positive examples and $\mathbf{N}_i(t)^-$ the negative ones. Without loss of generality, assume $e_i$ is positive, an example $e_i$ is safe if and only if 

$$\sum_{p \in \mathbf{N}_i(t)^+} \mathcal{A}(e_p) > \sum_{q \in \mathbf{N}_i(t)^-} \mathcal{A}(e_q) + \delta$$

\noindent where $k = \sqrt{b}$ and $b$ is the batch size and $\delta$ is the safety margin. $\mathbf{N}_i(t)^+$ and $\mathbf{N}_i(t)^-$ are computed by using the learned embedding at iteration t with the $\ell_2$ distance function.

\end{defi}

Inspired by the concept of borderline in SMOTE \cite{han2005borderline}, we consider a positive (negative) example is safe if the sum of assurance of its \emph{k} nearest positive (negative) neighbors are larger than the one from its negative (positive) neighbors. Please note that neighbors are defined in the embedding space and will dynamically change along with the model training. When constructing the \emph{n}-tuplets, we conduct the safety-aware sampling by giving safe examples higher probabilities compared to unsafe examples.

\subsubsection{Robust Anchor Generation}
\label{sec:anchor}

Anchors play one of the most important roles in the \emph{n}-tuplet based model learning. When learning from \emph{n}-tuplets, both the positive example and multiple negative examples are compared with the corresponding anchor in each \emph{n}-tuplet. The result of model learning highly relies on the quality of those anchors and any ambiguous anchor will lead to a suboptimal solution. Therefore, to reduce such inferior effect, we develop a robust anchor generation approach that artificially creates a batch-level ``gold standard'' anchor by summarizing all the anchors' information by their label assurance scores, i.e.,

\begin{defi}{(\textsc{Robust Anchor})} \label{def:anchor} Let $c_1, \cdots, c_m$ be the example indices of selected anchors and m be the total number of n-tuplets we generated within each training batch. The batch-level robust anchor is computed by:

$$e_r^* = \sum_{j = 1}^m \mathcal{A}(e^*_{c_j}) e^*_{c_j}$$

\end{defi}

After creating the robust anchor $e_r^*$, we replace all the original anchors from \emph{n}-tuplets within this batch with $e_r^*$. The robust anchor $e_r^*$ is more closer to the center of the cluster formed by the highly consistent examples and the influence of ambiguous anchors is significantly reduced. It is worth noting that the calculation of the robust anchor is easy to implement by adding a robust anchor generation layer in the SRL network, depicted in Figure \ref{fig:model}.

\subsubsection{SRL Network}
\label{sec:representation}

Inspired by the discriminative training approaches widely used in information retrieval \cite{huang2013learning,palangi2016deep} and natural language processing \cite{dos2014deep}, we present a supervised training approach to learning the representation network by maximizing the conditional likelihood of retrieving positive example $e^+$ given our robust anchor $e_r^*$ from the corresponding \emph{n}-tuplet with the rest $n-2$ negative examples. Similar to \cite{xu2019learning}, we design a weight-sharing deep neural network (DNN) for each example within the \emph{n}-tuplets.

Formally, given an embedding network parameterized by $\Theta$, let $\mathcal{F}_\Theta(e_i)$ be the learned representation of example $e_i$, we compute the posterior probability of $e^{+}$ given $e_r^*$ through a softmax function, i.e.,

$$p \big(e^{+}|e_r^* \big) = \frac{\exp \Big(\eta \cdot \mathcal{A}(e^+) \cdot \mathcal{R}(\mathcal{F}_\Theta(e^+), \mathcal{F}_\Theta(e^*_r))\Big)}{\sum\limits_{e_j \in \mathcal{T}} \exp \Big(\eta \cdot \mathcal{A}(e_j) \cdot \mathcal{R}(\mathcal{F}_\Theta(e_j), \mathcal{F}_\Theta(e^*_r))\Big)}$$

\noindent where $C$ and $\eta$ are smoothing hyper-parameters in the softmax function, which are set empirically on a held-out data set in our experiments. $\mathcal{R}(\cdot, \cdot)$ is the similarity function, i.e.,

$$\mathcal{R}(\mathcal{F}_\Theta(e_1), \mathcal{F}_\Theta(e_2)) \myeq C - \| \mathcal{F}_\Theta(e_1) - \mathcal{F}_\Theta(e_2) \|_{\ell_2}$$

To maximize the posterior, we would like to maximize the relevance between two positive embeddings $\mathcal{F}_\Theta(e^*_r)$ and $\mathcal{F}_\Theta(e^+)$, in the meanwhile, minimize the relevance between the robust anchor embedding $\mathcal{F}_\Theta(e^*_r)$ and all the other negative embeddings, i.e., $\{\mathcal{F}_\Theta(e^-_j)\}_{j=1}^{n-2}$. As distance is proportional to the inverse of relevance, similar data examples are pulled closer while dissimilar examples are pushed away in the embedding space.

Hence, given a collection of \emph{n}-tuplets, we optimize parameters of the embedding network by maximizing the sum of log conditional likelihood of finding a positive example $e^+$ given the robust anchor $e_r^*$ from \emph{n}-tuplet $\mathbf{T}$, i.e.,

$$\mathcal{L}(\Theta) = - \sum \log p \big(e^{+}|e_r^* \big)$$

Since $\mathcal{L}(\Theta)$ is differentiable with respect to $\Theta$, we use a gradient-based optimization approach to train the embedding network.

\subsubsection{\emph{N}-tuplet Sampling Network}
\label{sec:sampling}

By constructing the training \emph{n}-tuplets with safety-aware sampling and robust anchors, we are able to get quadratic or cubic training sample size compared to the original data set. On the one hand, we provide the embedding network sufficient training data and avoid the overfitting problem. On the other hand, the training process may become extremely long and may not guarantee optimal performance. Therefore, we explicitly design a sampling network to adaptively select effective training examples from a massive number of \emph{n}-tuplets.

Here we design our sampling network by \emph{n} parameter-sharing DNNs with multi-layer fully-connected projections. Every example in the \emph{n}-tuplet is passed to its corresponding DNN to get its non-linear embedding. These \emph{n} embeddings are concatenated and used as features to optimize the ``hardness'' score prediction. In general, the ``hardness'' scores can be calculated by any real-valued function. Here, we choose to use the training loss from the representation learning network as our surrogates of ``hardness'' scores. We use the square loss as the objective loss function of the sampling network.

\subsection{Joint Learning Paradigm}
\label{sec:joint}

We jointly optimize the SRL network and the \emph{n}-tuplet sampling network in an iterative manner, which is described as the following repeated steps:

\begin{itemize}
\item[Step 1:] The sampling network scores every $\mathbf{T}$ in the generated \emph{n}-tuplet collection, which is constructed by using safety-aware sampling and robust anchor generation. Those \emph{n}-tuplets with higher ``hardness'' scores are selected into set $\mathbf{H}$. We set the proportion of selecting ``hard'' \emph{n}-tuplets from all \emph{n}-tuplets to be 1/3 in our experiments.

\item[Step 2:] The representation network uses $\mathbf{H}$ for its parameter optimization. It passes the training loss $\mathcal{L}$ to the \emph{n}-tuple sampling network through forward-propagation. 

\item[Step 3:] The sampling network fine-tunes its parameters by utilizing the training loss $\mathcal{L}$ from the SRL network.
\end{itemize}

\section{Experiments}
\label{sec:experiment}
Experiments are conducted on both synthetic and real-world data sets from different domains. We would also like to note that the hyperparameters used in our methods are selected (in all experiments) by the internal cross-validation approach while optimizing models' predictive performances. We report accuracy and AUC scores to comprehensively evaluate the performance of our proposed method.  In the following, the proposed method is referred to as \emph{NeuCrowd}.

\subsection{Data}

\subsubsection{Synthetic Data} To get a good understanding of our approach, we first test it on a synthetic data set, i.e., \emph{Syn}. Here we use the same simulation approaches as Guyon et al. used in the NIPS 2003 variable selection task for generating synthetic samples \cite{guyon2004result}. Briefly, we create 4 clusters of points normally distributed (std=1) about vertices of a multi-dimensional hypercube with sides of length 2 and assign 2 clusters to each class (positive or negative) and sample features from these Gaussian distributions. Detailed can be found in \emph{scikit-learn.org} docs\footnote{\url{https://scikit-learn.org/stable/modules/generated/sklearn.datasets.make_classification.html}}. To generate crowdsourced labels, we consider the setting of multiple crowd workers, i.e., 7 workers. Here, crowdsourced labels are simulated by assigning a mislabeling probability for each worker. The mislabeling probability is obtained from a truncated normal distribution with an upper bound of 0.5, a lower bound of 0.01, a mean of 0.1, and a standard deviation of 0.1. 

\subsubsection{Pre-K Children Speech Data} We test our framework on a data set of a speech contest of children who prepare for primary school, i.e., \emph{Pre-K}. The contest examines the ability to address speech in front of a public audience. Each example is a 1-min video with binary labels indicating speech fluency (1 = \emph{fluent} and 0 = \emph{not fluent}). We extract both the linguistic features and acoustic features from the videos\footnote{Acoustic features are extracted by using OpenSmile, i.e., \url{https://www.audeering.com/opensmile/}}.

\subsubsection{Hotel Review Data} We use hotel comments, i.e., \emph{hotel}, as a benchmark data set. The data is collected from a third-party rating website. Each comment has a binary label indicating its positive or negative opinion (1 = \emph{positive} and 0 = \emph{negative}). The goal is to learn the language embedding to distinguish positive and negative comments.

\subsubsection{Vocal Emotion Data} We experiment with a vocal emotion data set (1 = \emph{affectionate} and 0 = \emph{plain}) to demonstrate the performance of the proposed framework, i.e., \emph{Emotion}. The emotion samples are the audio tracks obtained from students' free talks when studying the Chinese language. We extract the acoustic features from audio tracks$^5$.

\subsubsection{Data Statistics} 

We summarize the crowdsourcing settings and data statistics in Table \ref{tab:data}. In Table \ref{tab:data}, \emph{kappa} represents the value of Fleiss' kappa measurement \cite{fleiss1971measuring}, which is a statistical measure for assessing the reliability of agreement between a fixed number of raters when assigning categorical ratings to a number of items or classifying items. \emph{class ratio} denotes the class label ratio that is computed by the number of positive samples divided by the number of total samples. It should be noted that, for real-world data sets i.e., \emph{Pre-K}, \emph{Hotel}, and \emph{Emotion}, samples labeled by crowdsourcing workers are randomly split into training sets and validation sets with proportion of 80\% and 20\%, respectively. However, test samples are labeled by experts in order to be accurate.

\begin{table}
	\centering
	\caption{\label{tab:data}Data sets statistics.}
	\begin{tabular}{@{}lcccc@{}} \toprule
		& Syn & Pre-K & Hotel & Emotion \\ \midrule
		\# of annotators  & 7  &  11 &  7 &  5 \\ 
		\# of features  & 1200  & 1632  &  300 &  1582 \\ 
		\# of training samples   & 800  &  950 &   447 &  1942\\
		\# of validation samples  & 200  & 237 &  112 & 485\\
		\# of test samples  & 500  & 300 &  140 & 1500 \\
		kappa  & 0.52  & 0.60 &  0.80 & 0.47\\
		class ratio  & 0.50  & 0.65 &  0.43 & 0.34\\ \bottomrule
	\end{tabular}
\end{table}

\subsection{Baselines}
We carefully choose three groups of state-of-the-art methods as our baselines to comprehensively assess the effectiveness of the proposed framework.

\subsubsection{Group 1: True Label Inference from Crowdsourcing} The first group contains methods inferring true labels from crowdsourced labels. They are listed as follows:

\begin{itemize}
\item Logistic regression with every pair (instance, label) provided by each crowd worker as a separate example. Note that this amounts to using a soft probabilistic estimate of the actual ground truth to learn the classifier \cite{raykar2010learning}, i.e., \emph{SoftProb}. 

\item Logistic regression with GLAD labels \cite{whitehill2009whose}, i.e., \emph{GLAD}. GLAD jointly infers the true labels, worker's expertise, and the difficulty of each data instance.

\item Logistic regression with labels inferred by expectation-maximization with an initial estimate of worker confusion matrix by spectral methods \cite{zhang2014spectral}, i.e., \emph{SC}.

\item  Logistic regression with EBCC labels \cite{li2019exploiting}, i.e., \emph{EBCC}. EBCC captures worker correlation by modeling true classes as mixtures of subtypes, and in each subtype the correlation of workers varies.

\end{itemize}

\subsubsection{Group 2: SRL with Limited Labels} The second group includes SRL methods designed for limited labels. They are listed as follows:

\begin{itemize}
\item Contrastive Loss \cite{koch2015siamese}, i.e., \emph{Contrastive}. We train a Siamese network that learns an embedding with pairs of examples to minimize distance between intra-class instances.
\item Triplet networks with semi-hard example mining \cite{schroff2015facenet}, i.e., \emph{TripleSemi}. The triplet with the smallest distance between anchor and negative example in the embedding space is chosen.
\item Triplet networks with lifted structured loss \cite{oh2016deep}, i.e., \emph{Lifted}. Lifted structured loss is based on all the pairwise edges among positive and negative pairs of samples, which fully explores the relations of instances.
\item Triplet networks with center Loss \cite{he2018triplet}, i.e., \emph{Center}. Distance between each instance and the center (not weighted by vote confidence) is learned for each category, with the goal to minimize intra-class variations and maximize inter-class distances at the same time.

\item Learning with noisy labels by leveraging semi-supervised learning techniques \cite{li2019dividemix}, i.e., \emph{DivideMix}. The DivideMix models the per-sample loss distribution with a mixture model and trains the model on both the labeled and unlabeled data in a semi-supervised manner, which can be viewed as an extension of MixMatch proposed by  \cite{berthelot2019mixmatch}.

\item Learning an end-to-end DNN directly from the noisy labels of multiple annotators by using a general-purpose crowd layer \cite{rodrigues2018deep}, i.e., \emph{CrowdLayer}. It jointly learns the parameters of the network and the reliabilities of the annotators.

\item Loss correction with unsupervised label noise modeling \cite{arazo2019unsupervised}, i.e., \emph{LC}. It uses a two-component beta mixture model as an unsupervised generative model of sample loss values during training and corrects the loss by relying on the network prediction.

\end{itemize}

\subsubsection{Group 3: Two-stage Models by Combining Group 1 and Group 2} \textbf{Group 3} contains methods combining baselines from \textbf{Group 1} and \textbf{Group 2}. They solve the problems of the limited and inconsistent labels in two stages. Due to the page limit, we only combine the best approach in \textbf{Group 1} (\emph{SC}) with methods in \textbf{Group 2}. Please note that because the \emph{CrowdLayer} directly model each worker's annotation, it cannot be combined with methods in \textbf{Group 1}.

Please note that in this work, we deal with a more practical and realistic scenario where all labeling efforts are done in third-party annotation factories or crowdsourcing platforms. The majority of such paid services don't offer the option that pre-locking a fixed number of workers for each annotation task since (1) workers are usually part-time and unstable; and (2) such labeling resource pre-allocation reduces the overall annotation throughput. Therefore, we treat the crowdsourcing services as black boxes and we focus on improving the SRL performance without worker identities constraints.

\subsection{Implementation Details}

Experimental codes are implemented in Tensorflow 1.8\footnote{https://www.tensorflow.org/} and executed on a server with Ubuntu 14.04 LTS and a GTX 1080 Ti GPU. As suggested in \citeasnoun{xu2019learning}, we set \emph{n} to 5 for all the following experiments. We use a weight-sharing deep neural network with 2 fully-connected layers as the representation learning network and the sampling network. We set the dropout rate to 0.2. We initialize the network weights with a normal distribution initializer. We use Adadelta as our optimizer \cite{zeiler2012adadelta}. The learning rate for both embedding network and sampling network is set to 1e-3. Sizes of each layer and scale of $\ell_2$ regularization are hyper-parameters that are set by grid searching with cross-validation. Downstream logistic regression classifier is trained with the inverse of $\ell_2$ regularization strength \emph{C} as the only hyper-parameter ranging from 1e-4 to 1e4.

\subsection{Performance Comparison}

From Table \ref{tab:predict}, we make the following observations to compare performance of existing methods and \emph{NeuCrowd}:

\begin{itemize}
\item Methods in \textbf{Group 3}, which combine \emph{SC} with methods in \textbf{Group 2} to solve the problem of limited data and crowdsourced label inconsistencies at the same time, outperform baseline methods in \textbf{Group 2} on most of the data sets, which suggests that it's necessary to get rid of noises when using crowdsourced labels. In fact, simply training embeddings with majority-voted labels results in inferiority compared with classic methods.

\item Our framework performs better than networks trained with fixed \emph{SC} estimated labels in \textbf{Group 3}. Comparing \emph{NeuCrowd} and \emph{TripleSemi+SC}, \emph{DivideMix+SC}, \emph{LC+SC}, our approach utilizes the safety-aware sampling to only select good-quality examples into \emph{n}-tuples, which help the model get rid of ambiguous examples. When comparing \emph{NeuCrowd} to \emph{Lifted+SC} and \emph{Centered+SC}, instead of randomly selected ambiguous anchors, our framework makes full use of the assurance-aware anchors, which are more robust in crowdsourcing setting. Compared to \emph{Contrastive+SC} and \emph{Triple+SC}, our approach tries to learn the relations within multiple negative examples within each \emph{n}-tuple, which is more effective.
\end{itemize}

\begin{table*}
	\centering
	\small
	\caption{Prediction accuracy and AUC scores on both synthetic and real-world data sets. ``-'' represents the algorithm never converges. Paired t-tests are conducted to examine whether the \emph{NeuCrowd} has statistically higher accuracy than the compared methods, and ``*''  represents the significance at the level of 0.05.}
	\resizebox{1.0\textwidth}{!}{\begin{tabular}{r*{11}{l}}
		\toprule
		&& \multicolumn{2}{c}{Syn}  & \multicolumn{2}{c}{Pre-K} & \multicolumn{2}{c}{Hotel} & \multicolumn{2}{c}{Emotion} \\
		\cmidrule(lr){3-4} \cmidrule(lr){5-6} \cmidrule(lr){7-8} \cmidrule(lr){9-10}
		Method & Group & Acc & AUC & Acc & AUC & Acc & AUC & Acc & AUC\\
		\midrule
		SoftProb           	& group 1 & 0.666 & 0.720 & - & - & 0.850 & 0.921 & 0.819$^*$ & 0.917 \\
		GLAD                 & group 1 & 0.614$^*$ & 0.664 & 0.827$^*$ & \textbf{0.917} & 0.850 & 0.911 &  0.808$^*$ & 0.929 \\ 
		EBCC                 & group 1 & 0.640$^*$ & 0.681 & 0.787$^*$ & 0.885 & 0.864 & 0.912 &  0.766$^*$ & 0.854 \\ 
		SC                & group 1 & 0.654 & 0.716 & 0.833$^*$ & 0.920 & 0.857 & 0.918 &  0.830$^*$ & 0.934 \\ \midrule
		
		Contrastive 	  & group 2 & 0.580$^*$ & 0.590  & 0.820$^*$ & 0.875 & 0.850 & 0.908 &  0.792$^*$ & 0.840 \\
		TripleSemi        & group 2 & 0.638$^*$ & 0.678  & 0.654$^*$ & 0.631 & 0.771$^*$ & 0.795 & 0.844$^*$ & 0.892 \\
		Centered           & group 2 & 0.594$^*$ & 0.612  & 0.757$^*$ & 0.802 & 0.843 & 0.912 & 0.766$^*$ & 0.850 \\
		Lifted         	      & group 2 & 0.588$^*$ & 0.596  & 0.747$^*$ & 0.785 & 0.836$^*$ & 0.887 & 0.776$^*$ & 0.845 \\
		
		DivideMix   	  & group 2 & 0.578$^*$ & 0.631  & 0.677$^*$ & 0.704 & 0.865 & 0.853  & 0.666$^*$ & 0.749 \\ 
		CrowdLayer       & group 2 & 0.674 & 0.703  & 0.714$^*$ & 0.751 & 0.865 & 0.864 & 0.818$^*$ & 0.907 \\
		LC                     & group 2 & 0.560$^*$ & 0.611  & 0.653$^*$ & 0.628 & 0.786$^*$ & 0.812 & 0.632$^*$ & 0.742 \\
		
		 \midrule
		Contrastive+SC & group 3 & 0.596$^*$ & 0.603 & 0.827$^*$ & 0.862 & 0.850 & 0.906 & 0.836$^*$ & 0.940 \\
		TripleSemi+SC  & group 3 & 0.640$^*$ & 0.701 & 0.740$^*$ & 0.722 & 0.750$^*$ & 0.792 & 0.854$^*$ & 0.935 \\
		Centered+SC     & group 3 & 0.622$^*$ & 0.666 & 0.790$^*$ & 0.794 & 0.850 & 0.918 & 0.813$^*$ & 0.884 \\
		Lifted+SC          & group 3 & 0.588$^*$ & 0.605 & 0.750$^*$ & 0.790 & 0.857 & 0.911 & 0.830$^*$ & 0.910 \\
		
		DivideMix+SC   	  & group 3 & 0.614$^*$ & 0.677  & 0.680$^*$ & 0.708 & 0.857 & 0.854 & 0.685$^*$ & 0.791 \\
		LC+SC       & group 3 & 0.580$^*$ & 0.609  & 0.684$^*$ & 0.688 & 0.793$^*$ & 0.831 & 0.646$^*$ & 0.768 \\
		
		 \midrule
		NeuCrowd 	      & our   & \textbf{0.678} & \textbf{0.729} & \textbf{0.867} & 0.898 & \textbf{0.871} & \textbf{0.928} & \textbf{0.888} & \textbf{0.959} \\  
		\bottomrule
	\end{tabular}}
	\label{tab:predict}
\end{table*}

\subsection{Component Analysis}

\begin{figure}
	
	\begin{minipage}[c]{0.42\linewidth}
		\includegraphics[width=1\linewidth]{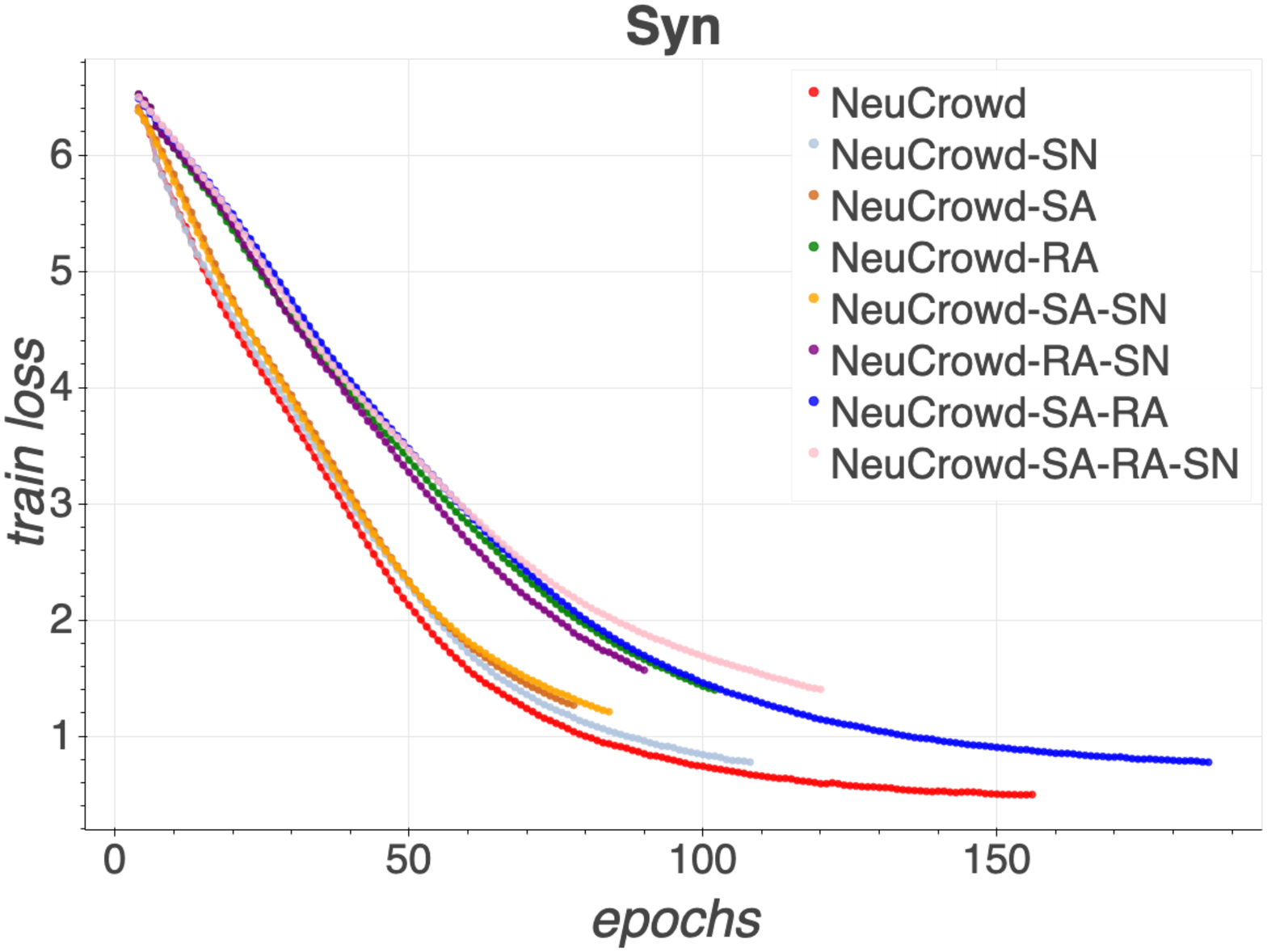}
		
		
	\end{minipage}
	\hspace{3ex}
	\begin{minipage}[c]{0.42\linewidth}
		
		\includegraphics[width=1\linewidth]{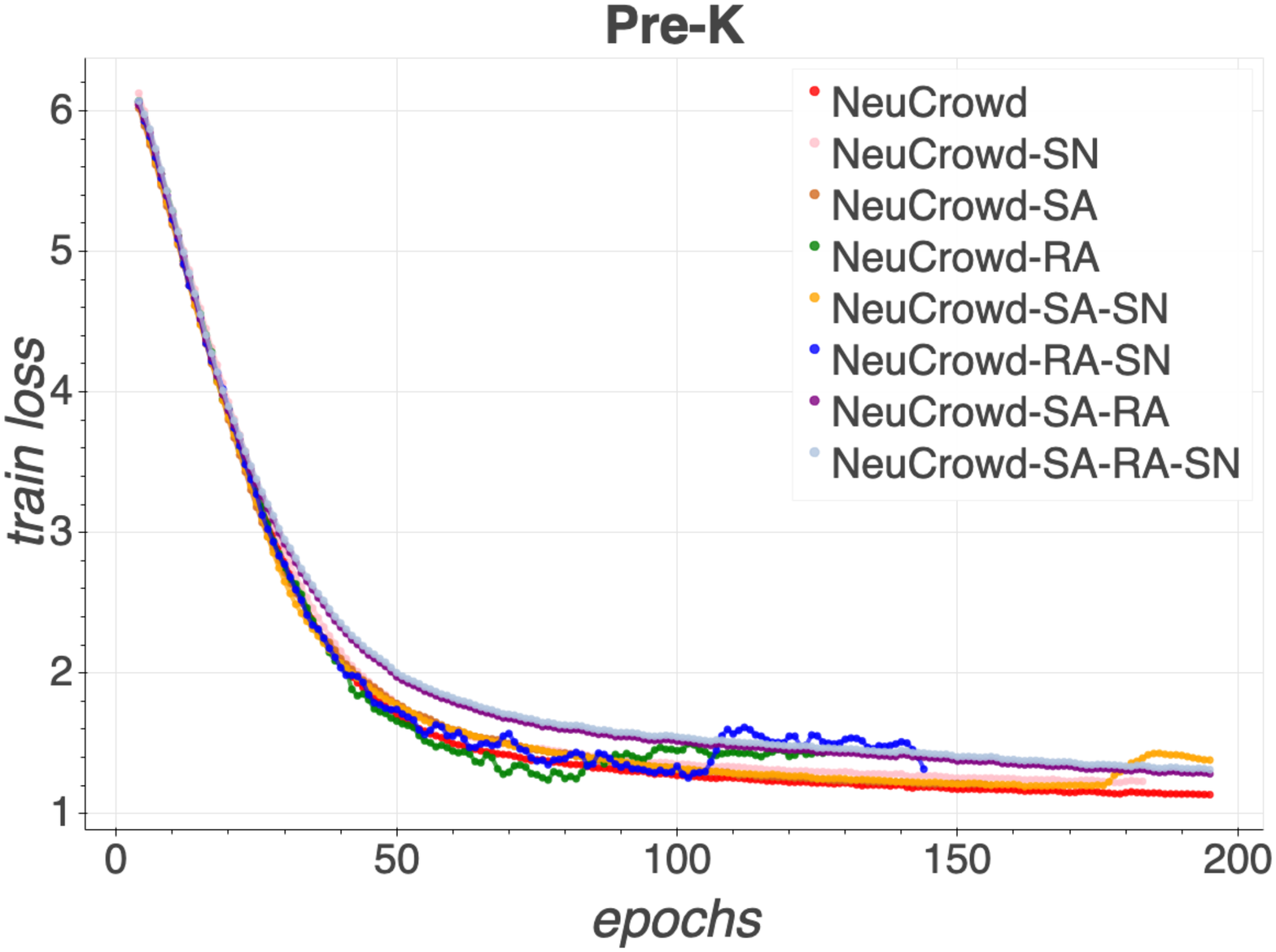}
		
		
	\end{minipage}
	\newline
	
	\begin{minipage}[c]{0.42\linewidth}
		\includegraphics[width=1\linewidth]{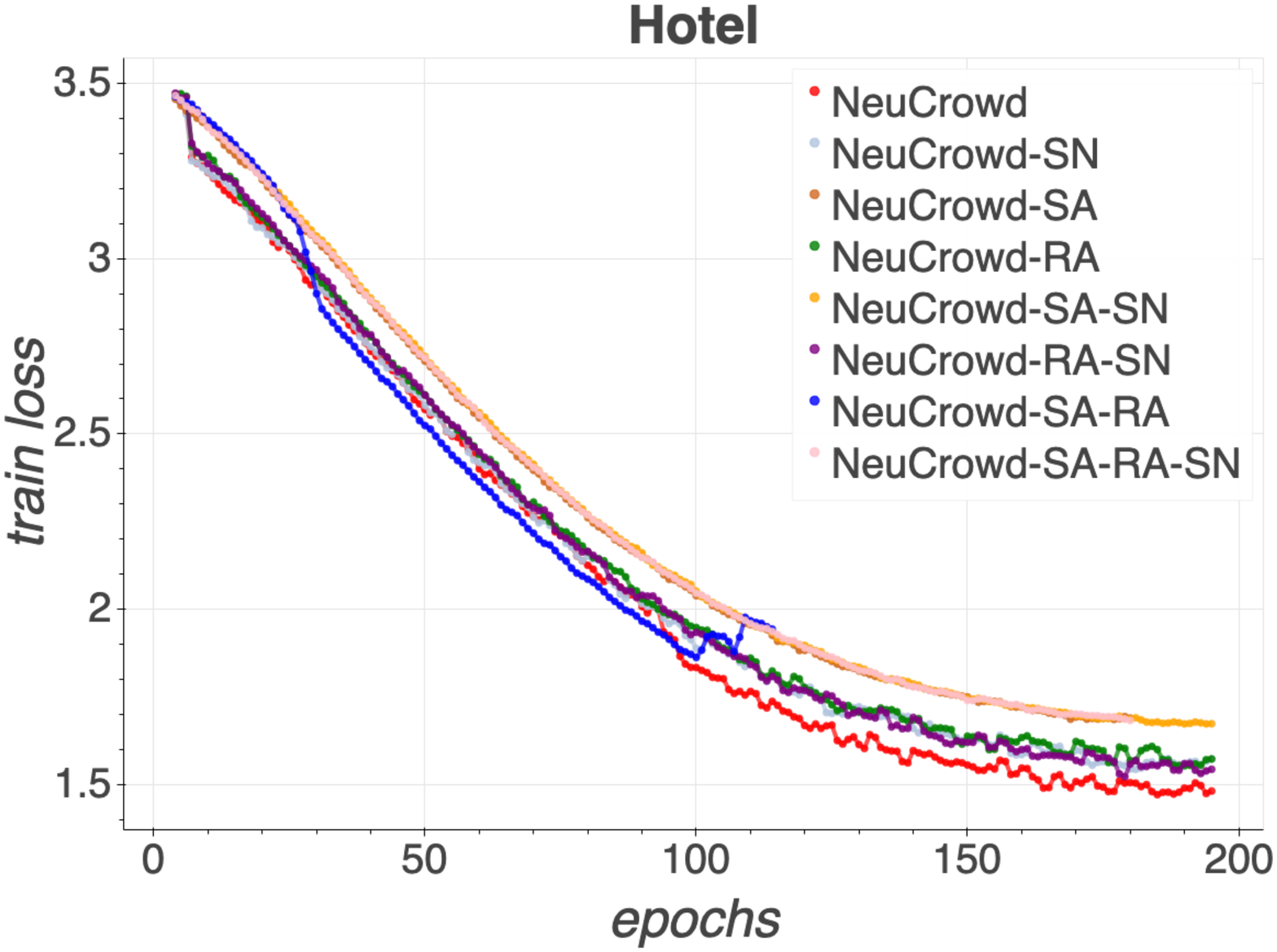}
		
		
	\end{minipage}
	\hspace{3ex}
	\begin{minipage}[c]{0.42\linewidth}
		
		\includegraphics[width=1\linewidth]{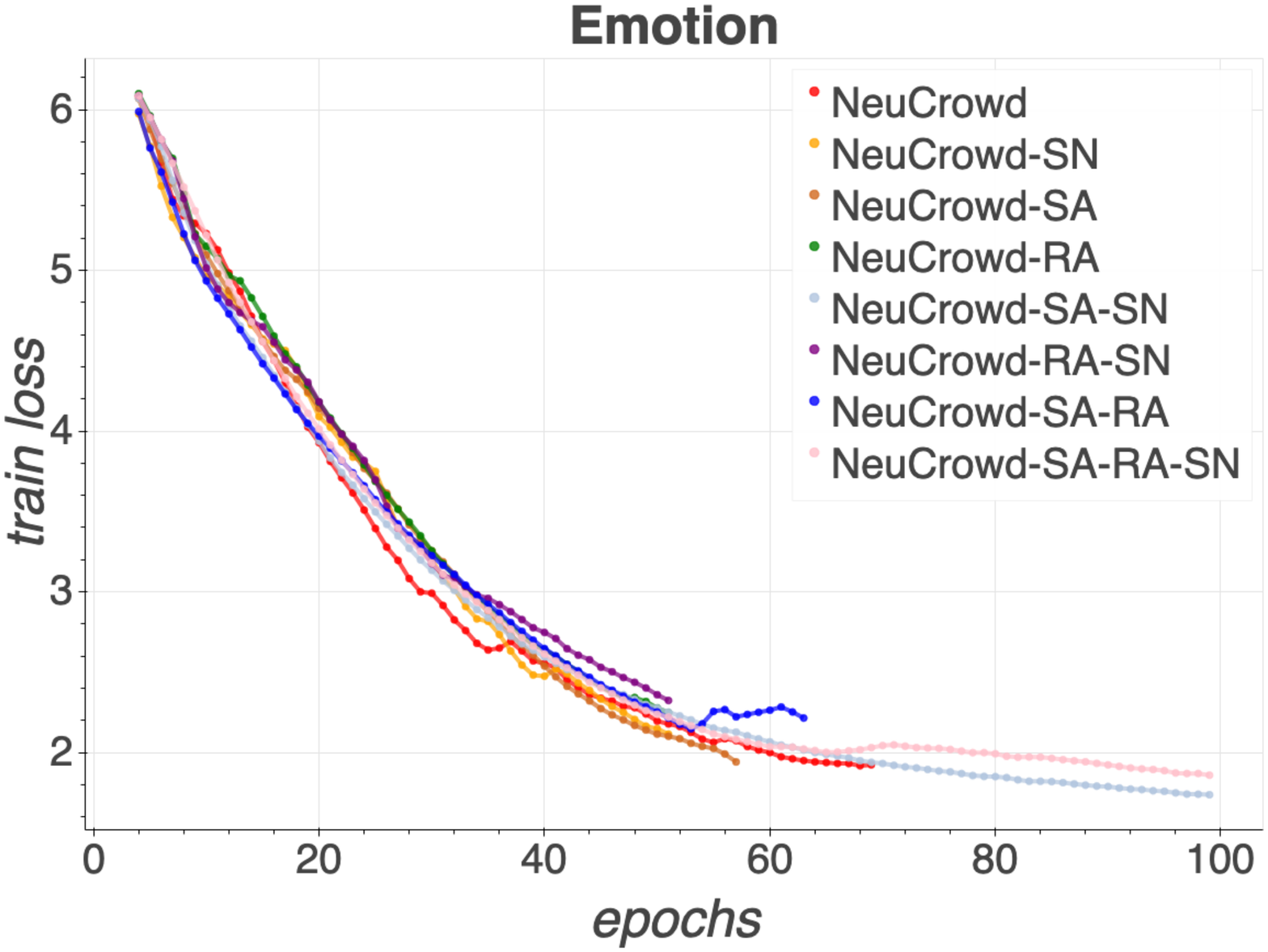}
		
		
	\end{minipage}
	
	\caption{The training loss of NeuCrowd with different components on different data sets.}
	\label{fig:train_loss}
\end{figure}

We systematically examine the effect of key components in the proposed framework. The full combinatorial model variants and their performance can be found in Table \ref{tab:component} and the changes of training loss are shown in Figure \ref{fig:train_loss}.

\begin{itemize}
\item NeuCrowd-SN: it eliminates the contribution of n-tuplet sampling network.
\item NeuCrowd-RA: it eliminates the contributions of robust anchors.
\item NeuCrowd-SA: it eliminates the contributions of safety-aware sampling.
\item NeuCrowd-RA-SN: it eliminates the contributions of both n-tuplet sampling network and robust anchors.
\item NeuCrowd-RA-SA: it eliminates the contributions of both safety-aware sampling and robust anchors.
\item NeuCrowd-SA-SN: it eliminates the contributions of both safety-aware sampling and n-tuplet sampling. 
\item NeuCrowd-SA-RA-SN: it eliminates the contributions of safety-aware sampling, robust anchors and \emph{n}-tuple sampling network and only the \emph{n}-tuple based representation learning model remains, which is equivalent to the \emph{RLL} framework proposed in \citeasnoun{xu2019learning}.  
\end{itemize}

\begin{table*}
	\centering
	\caption{Key component analysis of \emph{NeuCrowd} on both synthetic and real-world data sets.}

	\resizebox{1.0\textwidth}{!}{\begin{tabular}{r*{10}{c}}
		\toprule
		&\multicolumn{2}{c}{Syn}  & \multicolumn{2}{c}{Pre-K} & \multicolumn{2}{c}{Hotel} &\multicolumn{2}{c}{Emotion} & Time cost \\
		\cmidrule(lr){2-3} \cmidrule(lr){4-5} \cmidrule(lr){6-7} \cmidrule(lr){8-9}
		Method  & Acc & AUC & Acc & AUC & Acc & AUC & Acc & AUC &  on Emotion \\
		\midrule
		NeuCrowd-SA-RA-SN   & 0.650 & 0.712 & 0.807 & 0.867 & 0.771 & 0.838 & 0.794 & 0.885 & 1054\\
		NeuCrowd-SA-RA  & 0.665 & 0.713 & 0.807 & 0.883 & 0.829 & 0.921 & 0.776 & 0.840 & 1682\\ 
		
		NeuCrowd-RA-SN  & 0.668 & 0.705 & 0.817 & 0.850 & 0.857 & 0.917 & 0.841 & 0.892 & 1078\\ 
		NeuCrowd-SA-SN  & 0.642 & 0.708 & 0.790 & 0.823 & 0.793 & 0.852 & 0.840 & 0.920 & 1066 \\ 
		
		NeuCrowd-SN  & 0.670 & 0.718 & 0.860 & 0.919 & 0.857 & 0.916 & 0.857 & 0.914 & 1090\\ 
		
		NeuCrowd-SA  & 0.670 & 0.714 & 0.827 & 0.870 & 0.857 & 0.919 & 0.877 & 0.954 & 1683\\
		
		NeuCrowd-RA  & 0.665 & 0.704 & 0.814 & 0.853 & 0.843 & 0.916 & 0.873 & 0.935 & 1701\\ 
		
		NeuCrowd  & \textbf{0.678} & \textbf{0.729} & \textbf{0.867} &  \textbf{0.898} & \textbf{0.871} & \textbf{0.928} &  \textbf{0.888} & \textbf{0.959} & 1704\\ 
		\bottomrule
	\end{tabular}}
	\label{tab:component}
\end{table*}

When looking into the computational cost of the key components, we can find that the n-tuplet sampling network is trained by back propagation and it's expected to consume more time than the other two components, depending on the structure of the sampling network. In contrast, both robust anchor generation and safe example selection are computational-friendly. Robust anchor generation is performed with a time complexity of $\mathcal{O}(b)$ within an epoch, where $b$ is the batch size. As for safe example selection, distance calculation of vectors within a batch is conducted, which has mature solutions i.e., to be accelerated leveraging the Gram matrix. The training time costs (measured in seconds) of each combination above are recorded when training with the \emph{Emotion} data set, shown in Table \ref{tab:component}.

As we can see, our \emph{NeuCrowd} model outperforms all other variants in terms of prediction errors on all data sets. It is important to incorporate them together when building the end-to-end solutions of SRL from crowdsourced labels. Specifically, from Table \ref{tab:component} and Figure \ref{fig:train_loss}, we find the following results: (1) Without safety-aware sampling, robust anchors and \emph{n}-tuple sampling network, \emph{NeuCrowd-SA-RA-SN} has the worst performance. It may suffer from mislabeled examples and the corresponding learning process is not efficient. (2) By comparing NeuCrowd-RA with NeuCrowd, the role of robust anchors is illustrated that ambiguous anchors are replaced by the ``golden standard" generated anchor. (3) The fact that NeuCrowd outperforms NeuCrowd-SN shows that the process of representation learning can be improved by focusing on harder samples. (4) And by comparing NeuCrowd-SA with NeuCrowd, it's proved that samples with high quality can be explored by leveraging the learned representations. Finally, the full combination of three key components boosts the prediction performance.

\section{Conclusion}
\label{sec:conclusion}
We presented an SRL framework for learning embeddings from limited crowdsourced labels. Comparing with traditional SRL approaches, the advantages of our framework are: (1) it is able to learn effective embeddings from very limited data; (2) it automatically selects effective \emph{n}-tuplet training samples, which makes the training process more effective. Experimental results on both synthetic and real-world data sets demonstrated that our approach outperforms other state-of-the-art baselines in terms of accuracy and AUC scores.

\begin{acknowledgements}
This work was supported in part by National Key R\&D Program of China, under Grant No. 2020AAA0104500 and in part by Beijing Nova Program (Z201100006820068) from Beijing Municipal Science \& Technology Commission.
\end{acknowledgements}

\bibliographystyle{dcu}
\bibliography{kais2021.bib}

\section*{Author Biographies}
\leavevmode

\vbox{%
\begin{wrapfigure}{l}{80pt}
\vspace{-15pt}
\begin{center}
\includegraphics[width=0.18\textwidth]{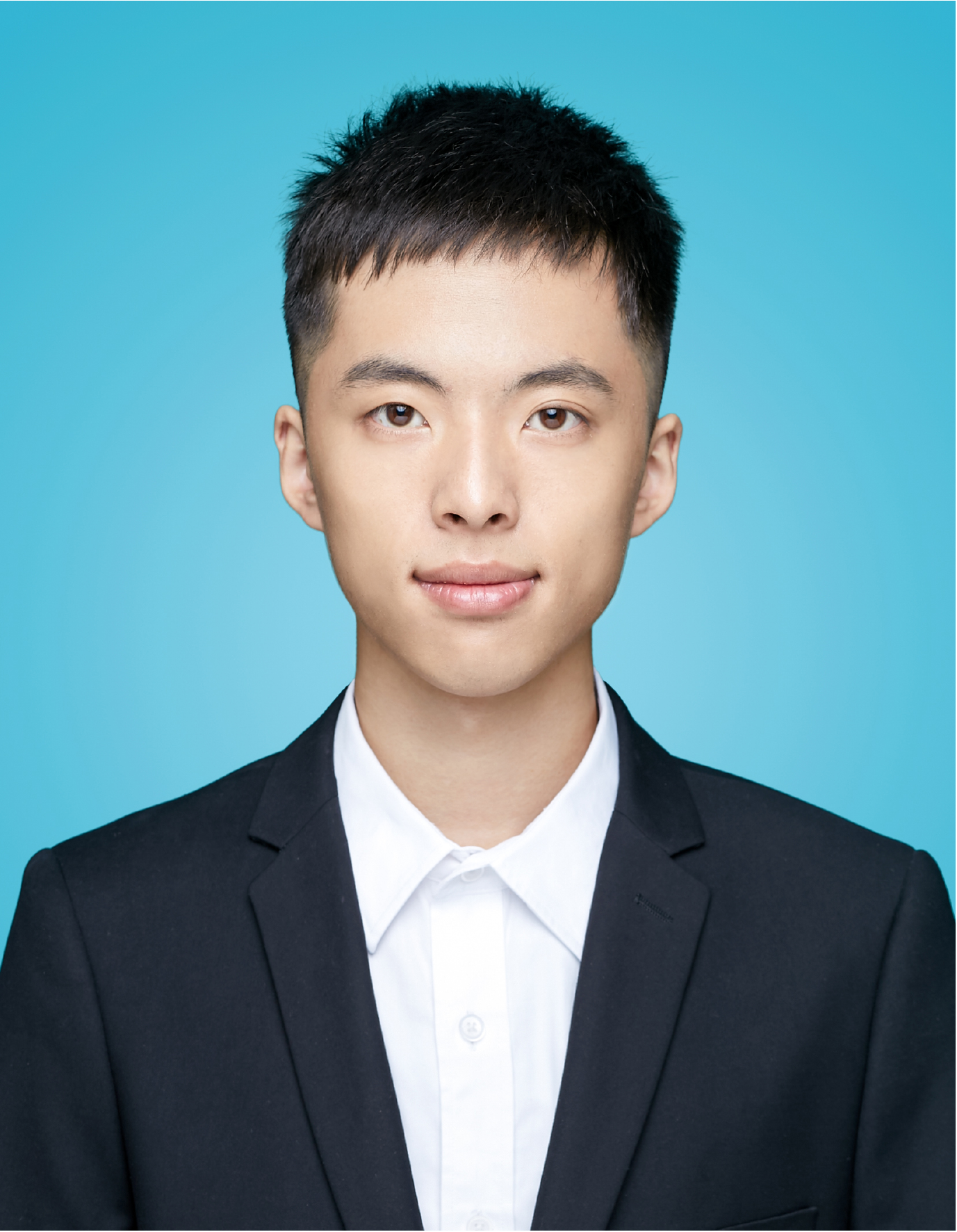}
\end{center}
\end{wrapfigure}
\noindent\small 
{\bf Yang Hao} is a machine learning engineer at TAL Education Group, China. He obtained his bachelor degree at Peking University. His research interests fall in areas of natural language processing, algorithms designed for crowdsourcing data, and AI for education.\vadjust{\vspace{50pt}}}

\vbox{%
\begin{wrapfigure}{l}{80pt}
\vspace{-15pt}
\begin{center}
\includegraphics[width=0.18\textwidth]{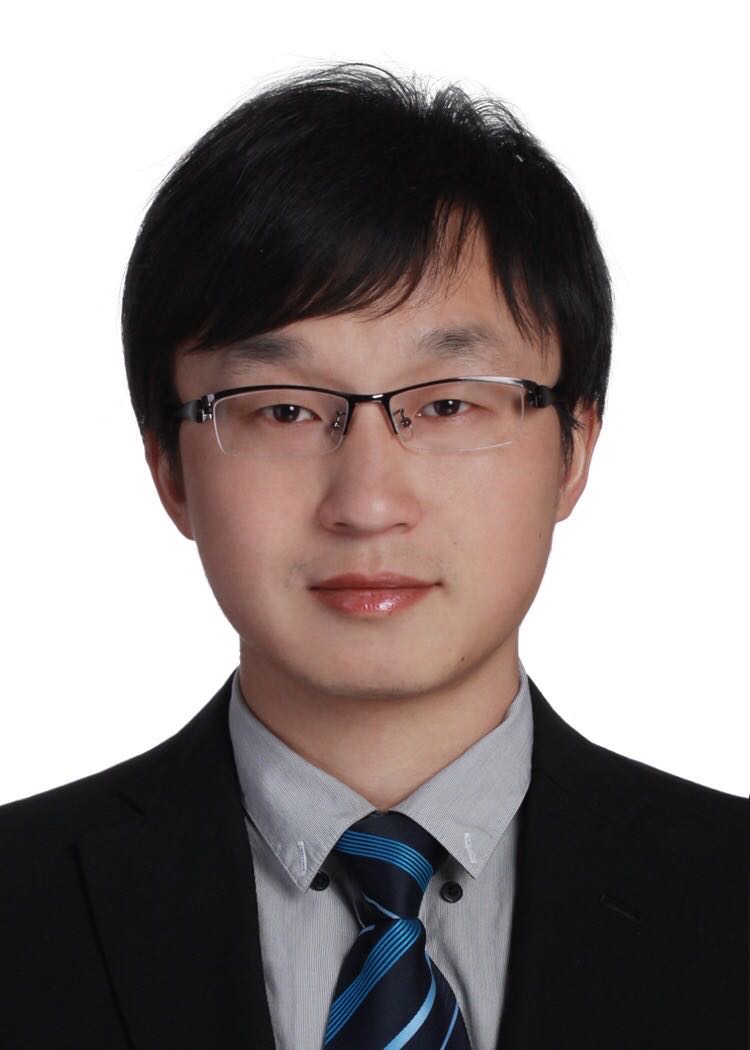}
\end{center}
\end{wrapfigure}
\noindent\small 
{\bf Wenbiao Ding} is a senior machine learning scientist at TAL Education Group, China. He has published several papers at top conference proceedings, such as ICDE, WWW, AIED, etc. He received his master's degree in computer science from the University of Science and Technology of China. Before joining TAL, Wenbiao was a senior research engineer at Sogou Inc. He worked on information retrieval, natural language processing and their applications in search engine systems and recommendation systems.\vadjust{\vspace{50pt}}}

\vbox{%
\begin{wrapfigure}{l}{80pt}
\vspace{-15pt}
\begin{center}
\includegraphics[width=0.18\textwidth]{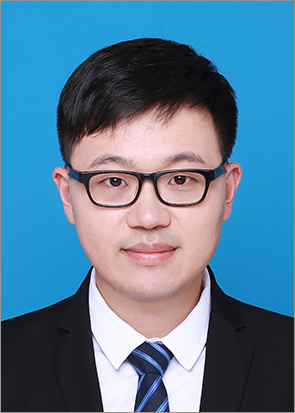}
\end{center}
\end{wrapfigure}
\noindent\small 
{\bf Zitao Liu} is currently the Head of Engineering, Xueersi 1 on 1 at TAL Education Group, China. His research is in the area of machine learning, and includes contributions in the areas of artificial intelligence in education, multimodal knowledge representation and user modeling. He has published his research in highly ranked conference proceedings and serves as the executive committee of the {\it International AI in Education Society} and top tier AI conference/workshop organizers/program committees. Before joining TAL, Zitao was a senior research scientist at Pinterest and received his Ph.D. degree in Computer Science from University of Pittsburgh.}

\correspond{Zitao Liu, TAL Education Group, Beijing, China. 
Email: liuzitao@tal.com}
\label{lastpage}
\end{document}